\documentclass{article}
\usepackage[utf8]{inputenc}
\usepackage{spconf}
\usepackage{float}
\usepackage{graphicx}
\usepackage{multirow}
\usepackage{graphicx}
\usepackage{hyperref}
\usepackage{subcaption}
\usepackage{graphicx}
\usepackage{algorithm}
\usepackage{algpseudocode}
\usepackage{amsmath}
\DeclareRobustCommand{\IEEEauthorrefmark}[1]{\smash{\textsuperscript{\footnotesize #1}}}
\usepackage{enumitem}
\usepackage{xcolor}
\newcommand{\ours}{VISION} 
\usepackage[skip=2pt]{parskip}
\title{UNIDIRECTIONAL BRAIN-COMPUTER INTERFACE: ARTIFICIAL NEURAL NETWORK ENCODING NATURAL IMAGES TO fMRI RESPONSE IN THE VISUAL CORTEX}

\name{%
\begin{tabular}{@{}c@{}}
Ruixing Liang,\IEEEauthorrefmark{1,2}$^{\star}$
Xiangyu Zhang,\IEEEauthorrefmark{1}$^{\star}$
Qiong Li,\IEEEauthorrefmark{3}
Lai Wei,\IEEEauthorrefmark{1,2}
Hexin Liu, \IEEEauthorrefmark{1}
Avisha Kumar \IEEEauthorrefmark{1,2}\\
Kelley M. Kempski Leadingham, \IEEEauthorrefmark{1,2}
Joshua Punnoose, \IEEEauthorrefmark{1,2}
Leibny Paola Garcia,\IEEEauthorrefmark{1}
Amir Manbachi\IEEEauthorrefmark{1,2}
\end{tabular}}
\address{
\IEEEauthorrefmark{1}Johns Hopkins University,
\IEEEauthorrefmark{2}Johns Hopkins Medicine,
\IEEEauthorrefmark{3}Pennsylvania State University
}

\begin{document}
\maketitle

\begingroup
\def\thefootnote{$\star$}
\footnotetext{Equal contribution}

\endgroup

\begin{abstract}
\vspace{1pt}
While significant advancements in artificial intelligence (AI) have catalyzed progress across various domains, its full potential in understanding visual perception remains underexplored. We propose an artificial neural network dubbed VISION, an acronym for “Visual Interface System for Imaging Output of Neural activity," to mimic the human brain and show how it can foster neuroscientific inquiries. Using visual and contextual inputs, this multimodal model predicts the brain's functional magnetic resonance imaging (fMRI) scan response to natural images. \ours ~successfully predicts human hemodynamic responses as fMRI voxel values to visual inputs with an accuracy exceeding state-of-the-art performance by 45\%.
We further probe the trained networks to reveal representational biases in different visual areas, generate experimentally testable hypotheses, and formulate an interpretable metric to associate these hypotheses with cortical functions. With both a model and evaluation metric, the cost and time burdens associated with designing and implementing functional analysis on the visual cortex could be reduced.
Our work suggests that the evolution of computational models may shed light on our fundamental understanding of the visual cortex and provide a viable approach toward reliable brain-machine interfaces. The source code can be found \footnote{Our open accessed repository can be found via \url{https://github.com/Rxliang/VISION}}.
\end{abstract}

\vspace{-3mm}

\section{Introduction}
\vspace{-2mm}
A fundamental pursuit in neuroscience is to uncover the neural basis of perceptions. Translating 1-dimensional auditory cues, which primarily captures amplitude changes over time into neural activity, is relatively well understood \cite{giordano_intermediate_2023}. On the other hand, visual stimuli's multifaceted 2D nature, encompassing attributes like color, texture, and depth, makes encoding more intricate \cite{khosla_characterizing_2022}.
Researchers frequently utilize functional magnetic resonance imaging (fMRI) to explore neural reactions triggered by visual stimuli. By measuring hemodynamic changes elicited by neural activity, fMRI illuminates the brain's approach to interpreting visual signals, serving as a potent modality in developing a brain-computer interface (BCI) aimed at modifying visual perception \cite{chen_survey_2014}.
\begin{figure}[!ht]
    \centering
    \includegraphics[width=0.46\textwidth, page=1]{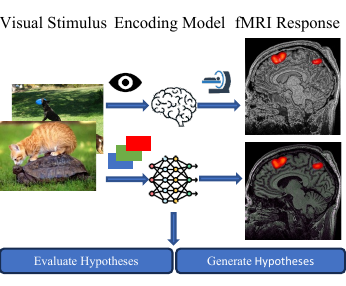}
    \caption{
    Overview of \ours, an Artificial Neural Network estimating functional Magnetic Resonance Imaging (fMRI) of the visual cortex response to visual stimuli. \ours ~ acts as a neural encoder that parallels the human visual cortex.
    }\vspace{-7mm}
    \label{fig:overview} 
\end{figure}
Consequently, there has been an increased emphasis on analyzing vision-related fMRI responses as subjects encounter various natural scenarios \cite{allen_massive_2022,chang_bold5000_2019}. 
Current standard continuous fMRI sessions, with a duration of 3-11 minutes and 3 mm resolution, come at a hefty cost of around \$1325 per session \cite{christina_what_2018}. To gather ample data, multiple fMRI sessions spread over a year are essential. The significant expenses have impeded obtaining high-quality data efficiently \cite{schneider_learnable_2023}. As a result, optimizing the use of existing datasets with minimal presuppositions to develop computational BCI models of human visual perception is paramount. By doing so, actionable hypotheses that guide subsequent research in a more focused and evidence-based direction could be generated, ensuring the efficient utilization of resources.
\begin{figure*}[ht]
    \centering
    \includegraphics[width=\textwidth]{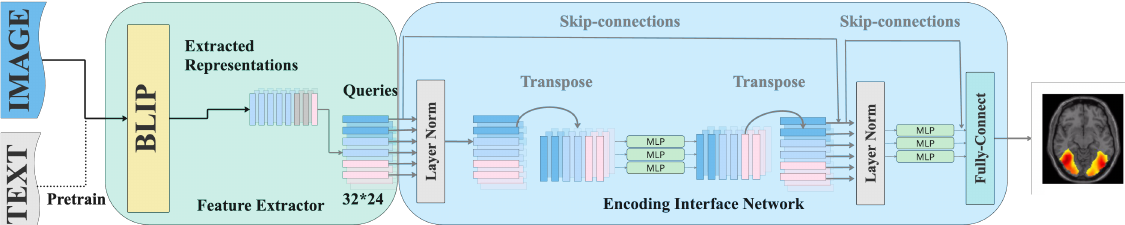}
    \caption{Overview of \ours ~model structure consisting of a feature extractor network and an encoding interface network}
    \label{fig:model}\vspace{-5mm}
\end{figure*}

Over the past few decades, deep learning (DL) has significantly transformed a myriad of scientific disciplines and industries. This transformation is largely attributed to the expanding scale of training datasets and the advancements in artificial neural networks (ANNs). Interestingly, given ANN's inherent parallel with human brain neural pathways and its performance, the study of artificial and biological intelligence is increasingly converging \cite{ghosh-dastidar_spiking_2009,hochreiter_long_1997,xuan2023new}.
Recognizing the potential of DL's nonlinearity and its robustness to noise, researchers have begun to leverage it in neuroscience \cite{khosla_characterizing_2022}. By treating neural responses, specifically fMRI data, as model inputs, several studies have employed generative ANN models to reconstruct visual stimuli, encompassing both static images and dynamic video streams \cite{schneider_learnable_2023,takagi_high-resolution_2022}. These DL models, often called “neural decoders" or “mind readers," provide a compelling avenue to decode brain activity. Conversely, there are DL models that predict neural responses based on visual stimuli, essentially acting as encoders. They offer valuable insights into the mechanism within cascading neural circuits in the human brain, revealing the intricacies of neural computation. Crucially, by integrating these two methodologies—decoding and encoding—a holistic loop for brain-computer interfaces can be established \cite{chen_survey_2014}.

\vspace{-1mm}
However, a significant limitation lies in these encoding models' scalability and interpretability. DL frameworks are often described as black boxes, leading them to be perceived as agnostic computational models with limited transparency \cite{gifford_algonauts_2023,li2023pqlm, li2023tatoo}.
To address these limitations, we introduce a novel multimodal (i.e., text and images) ANN model, \ours, ~with a scaled structure for predicting voxel-by-voxel fMRI responses. \ours’s accuracy has been evaluated for varying anatomical regions relevant to human visual processing and was found to align with hypotheses found in the neuroscience community \cite{popham_visual_2021,cordes_hierarchical_2002}. Additionally, we introduce a new approach for clear and quantifiable visual cortex functional analysis using class activation map (CAM)-based visualization and our purpose-built dataset \cite{wang_score-cam_2020}. This dataset is tailored to measure the model's attention across varied visual cues.
  

\vspace{-3mm}

\section{Methodology}
\vspace{-3mm}
\subsection{Multimodal Neural Encoding Model}
\vspace{-2mm}
As illustrated in \autoref{fig:model}, \ours ~consists of two fundamental building blocks: a multimodal feature extractor and a dense-channel encoding interface network.

\textbf{Feature Extractor}:
Inspired by recent work demonstrating that the visual cortex processes semantic contextual information in addition to visual input ~\cite{popham_visual_2021}, the state-of-the-art pre-training model has been adopted, BLIP~\cite{li_blip_2022}, as a feature extractor. The BLIP model consists of a vision transformer model~\cite{dosovitskiy_image_2020} and three transformer-based models with a similar structure to the BERT model~\cite{devlin_bert_2019}. Instead of using image-text pairs as inputs to the BLIP model, only images have been used to reduce computational complexity. The output of the BLIP feature is a high-dimensional feature containing textual information from the pre-training.

\vspace{-1mm}
\textbf{Encoding Interface Network}:
Given that the Multilayer Perceptron (MLP) draws inspiration from the structure of brain neurons, encoding interface network's design was grounded on the MLP model. MLP-Mixer has demonstrated meritorious performance across various computer vision tasks \cite{tolstikhin_mlp-mixer_2021}. Each MLP model consists of two fully-connected layers and one GELU activation function \cite{hendrycks_gaussian_2023}. For the MLP model to better understand the features from BLIP, we performed a series of processing steps on the BLIP features. First, the BLIP features are converted into a 32*24 two-dimensional matrix, calling it a query. The reconstruction is intended to replicate the complex, layered organization of the cerebral cortex. 197 queries are produced due to the properties of the BLIP feature, followed by processing in an MLP model for each query and combined as shown in \autoref{fig:model}.
\vspace{-6mm}
\subsection{Feature Space Visualization through Dimension Reduction}
\vspace{-2mm}
Dimension reduction plays a crucial role in visualization and as a preprocessing tool in deep learning due to the inherent challenges of high dimensionality \cite{maaten_visualizing_2008}. Specifically,  Uniform Manifold Approximation and Projection for Dimension Reduction (UMAP) has been employed to transform high-dimensional data into a compact representation, maintaining both local and global structures \cite{mcinnes_umap_2020}.  We extracted condensed features from the encoding interface network for visualization and paired each feature with augmented supercategories, enabling unsupervised visualization and evaluation of the model feature space.

\vspace{-3mm}
\subsection{Neuroscientific Hypotheses Testing and Generation via CAM Visualization}
\begin{figure}[!t]
    \centering
    \includegraphics[width=0.48\textwidth, page=1]{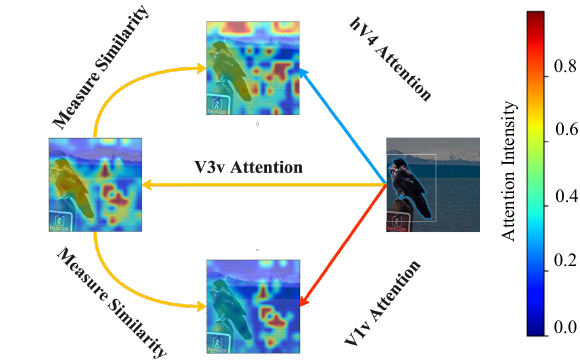}
    \caption{
    Illustration of ScoreCAM visualization to get the model's attention map of a specific region of interest in the visual cortex(i.e., hV4, V3v, and V1v)
    } \vspace{-5mm}
    \label{fig:cam} 
\end{figure}

\vspace{-2mm}
In this study, we employed ScoreCAM to elucidate the pixel-wise contribution from the visual image input towards each single voxel prediction, which can intuitively be interpreted as the model's attention map as illustrated in \autoref{fig:cam} \cite{wang_score-cam_2020}.

We hypothesize that the attention map of a visual cortex sub-region from the \ours ~model reflects the input image region that its biological counterpart would process. By evaluating the similarities and distinctions across regions in the visual cortex (\autoref{fig:cam}), we not only verify established neuroscientific hypotheses but also pave the way for formulating new hypotheses for regions whose focus is not known. Our primary emphasis has been testing hypotheses related to the higher visual centers, particularly those concerning object comprehension ~\cite{khosla_characterizing_2022}.
First, a group of images with one main object were selected. Following, Segment Anything (SAM) was used to segment and extract the main object from the image (e.g., a cat or dog in the center) \cite{kirillov_segment_2023}. Then, we overlapped the attention maps (from ScoreCAM) with the segmented objects to compute the dissimilarity of the overlapping regions between the different visual cortex regions, further quantified by Kullback–Leibler divergence. 
These attention maps can also be used to predict the role of varying visual cortex  regions in object comprehension by dividing the attention score of the primary object by the total attention score, as described in \autoref{camp}:
\begin{equation}
P_{f} = \frac{1}{n} \sum \frac{\mu_{AF}}{\mu_{OF}+\mu_{AF}} \label{camp},
\end{equation}
where $P_{f}$ is the probability of visual cortex groups belonging to a particular function $f$, $\mu_{AF}$ is the mean of attention distribution in some particular regions normalized with a particular area matching a function (e.g., object, edge), and $\mu_{OF}$ is mean of the attention out of the particular function. Using this process, the relationship between region characteristics of the visual image and neuronal activation can be mapped. It may spark new hypotheses on the role of varying visual cortex regions in visual processing. As this paper primarily focuses on visual comprehension, our analysis was limited to regions associated with the higher-order visual centers that control scene comprehension and object recognition\cite{khosla_characterizing_2022}. 
\vspace{-3mm}
\section{Experiments}
\vspace{-3mm}
\subsection{Dataset}
Our dataset comprised fMRI recordings from 8 participants as they viewed between 9,000 and 10,000 natural scenes \cite{allen_massive_2022}. We have integrated the Common Objects in Context (COCO) dataset source of the visual stimulus to augment this dataset by adding text descriptions. This produced image-fMRI pairs and image-caption-fMRI triads. Given the original human captions' verbosity, we devised an automatic labeling and cross-validation pipeline inspired by the OpenCLIP library, preparing our model's multimodal input \cite{cherti_reproducible_2022}.

Unlike existing studies that focus solely on the five major visual cortical regions of interest (ROIs) \cite{khosla_characterizing_2022}, we encompass the entirety of the visual cortex's anatomical structure for a total of 27 regions. Moreover, we have preprocessd resting state fMRI scans using the dataset's atlas, delivering robust functional connectivity analysis suitable for heuristic hypothesis input and cross-validation. Lastly, 100 images have been labeled semi-automatically, based on the current understanding of the visual cortex functions, to design our metric for functional analysis on different regions of the visual cortex \cite{kirillov_segment_2023}.
\vspace{-2pt}
\begin{figure*}[ht]
    \centering
    \includegraphics[width=\textwidth]{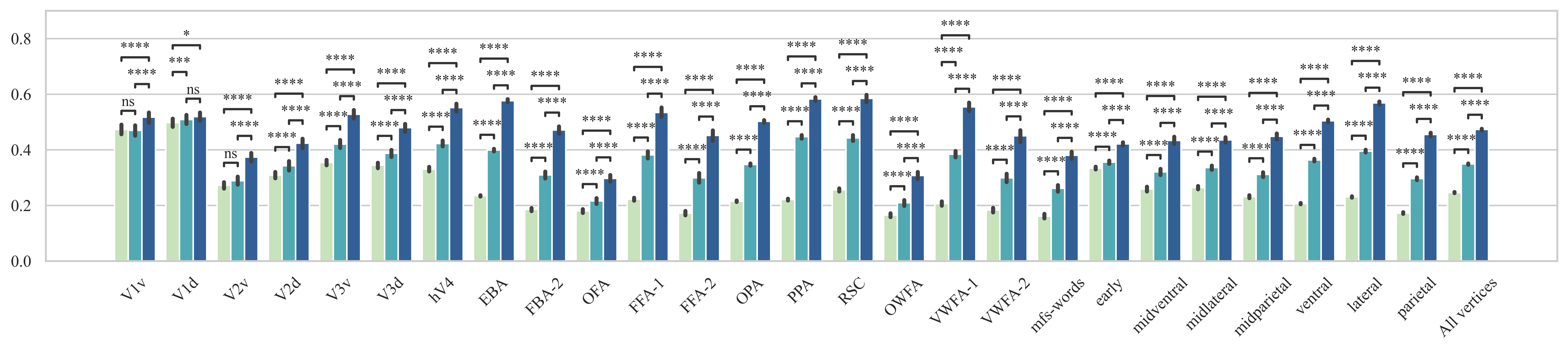}
    \caption{Prediction accuracy comparison with vanilla model by Allen et al. \cite{allen_massive_2022} and baseline model by Gifford et al.  \cite{gifford_algonauts_2023}. ns: no significance; *$p<0.05$; **$p<10^{-3}$; ***$p<10^{-4}$; ****$p<10^{-5}$}
    \label{fig:res}\vspace{-3mm}
\end{figure*}
\vspace{-4mm}
\subsection{Experimental Settings and Neural Architecture Search}
  80\% of the dataset from each participant has been allocated for training, and the remaining 20\% has been reserved for evaluation and testing. This approach was implemented to ensure model generalization across individuals. Each model was tasked with predicting a hemisphere in a participant's brain.

To evaluate prediction accuracy, noise-normalized accuracy is adopted. This entails the noise ceiling  \cite{lage-castellanos_methods_2019}, using \autoref{noise}:
\begin{equation}
\mathrm{NC}=100 \times \frac{\sigma_{\text {signal }}^2}{\sigma_{\text {signal }}^2+\sigma_{\text {noise }}^2}.
\label{noise}
\end{equation}

NC is the noise-ceiling value associated with each voxel, defined as the maximum percentage of variance in the voxel's responses contributed by signal ($\sigma_{\text{signal}}$) given the presence of estimated noise ($\sigma_{\text {noise}}$) 
provided by the dataset. Then accuracy is calculated according to \autoref{R} and \autoref{acc} \cite{gifford_algonauts_2023}.
\begin{align}
R_v & = \operatorname{corr}\left(G_v, P_v\right) \nonumber \\
& = \frac{\sum_t\left(G_{v, t}-\bar{G}_v\right)\left(P_{v, t}-\bar{P}_v\right)}{\sqrt{\sum_t\left(G_{v, t}-\bar{G}_v\right)^2 \sum_t\left(P_{v, t}-\bar{P}_v\right)^2}},
\label{R}
\end{align}
\vspace{-3mm}
\begin{equation}
\text { Accuracy }=\text { Median }\left\{\frac{R_1^2}{N C_1}, \ldots, \frac{R_v^2}{N C_v}\right\} \times 100,
\label{acc}
\end{equation}
where $G_v$ and $P_v$ represent the ground truth and predicted fMRI values of a voxel and its corresponding visual stimuli indexed, respectively, by $v$ and $t$.

Hyperparameter tuning and Neural Architecture Search (NAS) have been executed parallel to finetune each model. Both have followed the Tree-structured Parzen Estimator (TPE) paradigm \cite{bergstra_algorithms_2011}. We adopted Adam Optimizer with a finetuned initial learning rate for each model. The training environment was based on Pytorch distributed in an RTX3090 machine and an A100 machine randomly. 
\vspace{-3mm}
\subsection{Results}
\vspace{-1mm}
\subsubsection{Benchmark}
\vspace{-7pt}We have compared both the quantitative and qualitative outcomes of \ours ~models against two established baselines (i.e., vanilla and baseline model) \cite{allen_massive_2022,khosla_characterizing_2022,gifford_algonauts_2023}. Furthermore, our models exhibit superior performance when benchmarking against even unpublished research in some regions\cite{adeli_predicting_2023}. A summary of the quantitative comparisons can be found in \autoref{fig:res}. Pair-wise t-tests have been used to assess the significance of observed differences. Overall, for all vertex accuracy, \ours ~models significantly exceed baseline models. It is worth noting that improvements in the early visual cortex are minimal (i.e., V1, V2, V3). This observation aligns with our initial hypothesis: these regions do not process semantic information. Additionally, the accuracy improvements on the peripheral regions are more significant, aligning with the neuroscientific understanding that semantic information is processed on the periphery \cite{popham_visual_2021}. This evidence demonstrates the alignment of \ours~with recent visual cortex discoveries, warranting further investigation into the ability of this model to capture biological functionalities.
\vspace{-10pt}
\subsubsection{Feature Space}
\begin{figure}[!t]
    \centering
    \includegraphics[width=0.48\textwidth, page=1]{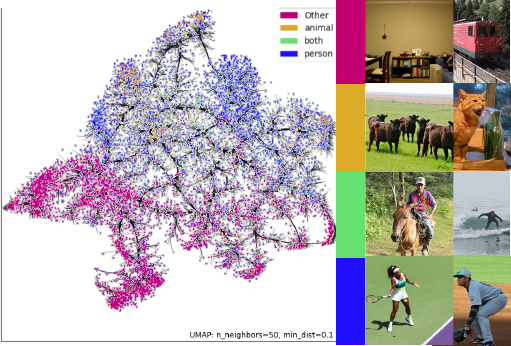}
    \caption{
    Feature map visualization with the global and local link bundling overlay (black lines) for varying image features (i.e., other, animal, person, and both [animal + person]). Sample images are displayed accordingly on the right.
    }
    \label{fig:vis} \vspace{-5mm}
\end{figure}

\vspace{-7pt}
As illustrated in \autoref{fig:vis},  response patterns within all participants' \ours ~models tend to group semantically. Typically, such distinct clustering, where identical categories are close-knit while being separated from dissimilar ones, emerges in networks that have been specifically trained and regularized to differentiate between these categories. Moreover, when compared with previous works, this clustering performance is better \cite{allen_massive_2022}. For the “other" category, its broader nature results in a broader distribution. The overlap labeled “both" signifies the presence of a person and an animal within a single image, and its placement within the respective person and animal clusters follows expectations. 
\vspace{-10pt}
\subsubsection{Hypotheses Testing on Functional Connectivity}
\vspace{-7pt}Following the traditional use of ScoreCam on BLIP'S vision encoder's last normalization layer, we have selected hV4, V3v, and V1v as testing regions. According to KL divergence, results show that the similarity of hV4 and V3v is 3.72 times more similar than hV4 and V1v, which could also be visually inferred in \autoref{fig:cam}. This is consistent with functional connectivity analysis derived from resting state fMRI and current findings \cite{cordes_hierarchical_2002}. This high similarity demonstrates a new way of deciphering the brain by first validating existing neuroscientific hypotheses with attention and then calculating the probability of the region's possible function using the attention ratio (\autoref{camp}). We test the object comprehension function of the hV4 region. Our results show that hV4 has a 59.5\% probability of being used for object comprehension. This further shows \ours ~model resemblance with the biological visual cortex. 

\vspace{-3mm}
\section{Conclusion}
\vspace{-3mm}
This paper introduces a transformer-based ANN model, termed \ours, designed to predict fMRI responses from visual stimuli. Utilizing a multimodal feature extractor, \ours ~processes visual cues and leverages pre-trained semantic information. This data then feeds into a dense-channel encoding interface network, significantly exceeding state-of-the-art accuracy. Evaluation of \ours ~ demonstrates the model's performance and interpretability and suggests its generalizability to traditional computer vision tasks. We further analyze the model's attention map, using quantifiable metrics to test existing theories. This approach illuminates the potential benefits of integrating large-scale ANN models in neuroscience. This synergy promises progress in BCIs and presents an exciting pathway to advance our fundamental understanding of the visual cortex.

\section{Acknowledgements}
The authors declare that there is no conflict of interest. A.M., K.K.L, A.K., and R.L. acknowledge funding support from the Defense Advanced Research Projects Agency (DARPA) Award (N660012024075). Q.L. acknowledges funding support from the National Institutes of Health (NIH) Award (R01NS085200). J.P. acknowledges funding support from NIH Research Training Award (5T32AR67708-8).

\bibliographystyle{IEEEbib}

\begin{footnotesize}
\bibliography{reference}
\end{footnotesize}

\end{document}